\documentclass[runningheads]{llncs}
\usepackage[T1]{fontenc}
\usepackage{graphicx}
\usepackage{cite}
\usepackage{amsmath,amssymb,amsfonts}
\usepackage{algorithmic}
\usepackage{graphicx}
\usepackage{textcomp}
\usepackage[table,xcdraw]{xcolor}
\usepackage{xcolor}
\usepackage{multirow}     
\usepackage{makecell}     
\usepackage[table]{xcolor} 
\usepackage{booktabs}    

\usepackage{float}

\usepackage{booktabs}
\usepackage{comment}
\usepackage{multirow}
\usepackage{array}
\usepackage[normalem]{ulem}
\usepackage{stfloats}
\usepackage{todonotes}
\newcommand{\hugo}[1]{\todo[inline,color=green!40]{#1 -- Hugo}}

\begin{document}
%
\title{Enhancing RAG-based Medical VQA with\\ Gated Multimodal Representations}
%
%
\author{Daniela Jimenez-Pano, Hugo Jair Escalante-Balderas and Delia Irazú Hernández-Farías}
\author{Daniela Jimenez-Pano~\inst{1}
\and
Hugo Jair Escalante\inst{1,2,3}
\and
Delia Irazú Hernández-Farías\inst{1}
}
%
%
\institute{$^1$LabTL - INAOE, Puebla, Mexico\\  
$^2$Institute for AAII, UTEP, El Paso, TX, USA\\ 
$^3$Department of Computer Science, UTEP, El Paso, TX, USA\\\email{danielapano@inaoep.mx}
}
%

%

\maketitle              
\begin{abstract}
Medical Vision-Language Models (Med-LVLMs) have shown strong performance in Medical Visual Question Answering (MVQA), yet they may still produce factually inconsistent responses. 
Retrieval-Augmented Generation (RAG) can mitigate this issue by incorporating external clinical evidence; however, retrieval is often guided by a single modality, potentially overlooking the joint semantics of medical images and clinical questions. 
We introduce a multimodal RAG framework that learns a joint query representation through a \textit{Gated Multimodal Unit}, enabling evidence retrieval to leverage information from both modalities while keeping the underlying Med-LVLM frozen.
We evaluate the proposed approach on benchmark datasets. 
Results show that the multimodal query representation consistently outperforms a reference Med-LVLM and remains competitive with state-of-the-art methods. 
Our  findings demonstrate the potential of multimodal query representations for more effective evidence retrieval in Medical VQA.
\end{abstract}

\section{Introduction}
Progress in Medical Large Vision-Language Models (Med-LVLMs) has enabled promising interactive diagnostic applications~\cite{xia2025mmedrag}. 
However, these models may still produce factually 
inconsistent responses, particularly in tasks such as Medical Visual Question Answering (Medical VQA), report generation, and clinical decision support, where generated outputs may not align with medical knowledge or reference answers. 
Retrieval-Augmented Generation (RAG) can mitigate this issue by incorporating external clinical evidence into the inference process. 
In this context, traditional RAG combines a parametric model with a non-parametric retriever, typically using Maximum Inner Product Search, to select the top-$k$ relevant documents from a dense knowledge base~\cite{lewis2020rag}. 
However, in multimodal  scenarios, 
retrieval is often guided by a single modality, even when the input includes both a medical image and a clinical question~\cite{xia2025mmedrag}. 
This may lead to misalignment of the evidence in the query, where the retrieved passages partially reflect the multimodal clinical context. 

We propose a multimodal query representation for retrieval in Medical VQA. 
Our approach incorporates a Gated Multimodal Unit (GMU)~\cite{arevalo2017gmu} to combine visual and textual features, 
producing a joint image-question embedding. The learned representation is
 used as the query-vector to retrieve the top-$k$ relevant textual passages from a vector database, while keeping the Med-LVLM frozen. 
By focusing on retrieval representation, our method aims to better align the multimodal input with the retrieved clinical evidence.  Experimental results on benchmark Medical VQA datasets show that the proposed multimodal retrieval representation boosts the performance of a reference Med-LVLM, and remains competitive with state-of-the-art models. 
Our contributions are as follows: 
\vspace{-0.2cm}
\begin{itemize}
\item Introducing a \textbf{novel methodology for learning multimodal query representations} in a RAG framework by integrating a GMU into the retrieval stage, 
explicitly modeling the interaction between the involved modalities. 
\item Proposing \textbf{four joint representation learning loss-function configurations} 
that emphasize different aspects of the available information. 
\item Performing \textbf{an extensive experimental evaluation} 
of the proposed methodology on benchmark Medical VQA datasets, showing its competitiveness in challenging scenarios. 
\end{itemize}

\subsection{Related work} \label{sec:related}
RULE~\cite{xia2024rule} and MMed-RAG~\cite{xia2025mmedrag} 
retrieve textual reports using mainly 
visual information. 
RULE uses a CLIP-based image-text retriever and a factuality risk-control strategy to select the number of contexts $k$; while MMed-RAG uses domain-aware retrievers and adaptive context selection. 
Both further apply preference tuning to reduce over-reliance on retrieved information.
FactMM-RAG~\cite{sun2025factaware} follows a similar visual-centric formulation for report generation: the approach employs MARVEL~\cite{zhou2024marvelunlockingmultimodalcapability} to represent medical images and reports in a shared embedding space. 
\emph{Thus, although its retriever is trained using paired visual and textual information, the query used at inference remains based on the visual modality.}
RAMM~\cite{yuan2023ramm} fuses the image-question representation with retrieved image-text pairs through a retrieval-attention module, and MOTOR~\cite{shaaban2025motor} uses grounded captions and optimal transport during re-ranking.
MIRA~\cite{MIRA} weights modalities dynamically over a multimodal index, and MM-RAG-MedVQA~\cite{shaaban2025towards} builds a query from the image and a generated caption, but leaves the question out of the retrieval representation, incorporating it only at generation time. 
\emph{Hence, multimodal fusion has been explored for processing 
evidence, but not as a 
joint 
representation to be used directly as the initial
retrieval query.}


\section{Proposed method} \label{sec:meth}
\begin{figure}[htb] 
    \centering
\includegraphics[width=0.8\textwidth]{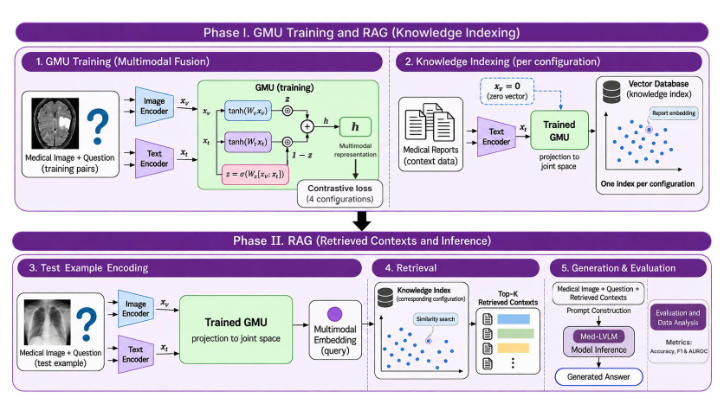} 
    \caption{Overview of the proposed  framework for Medical VQA. 
    }
    \label{fig:mvqageneral}
\end{figure}

The proposed framework is depicted in Figure~\ref{fig:mvqageneral}, it comprises five steps grouped into two stages. Each step is described in the remainder of this section. 
\paragraph{1. GMU Training (Multimodal Fusion).} To learn multimodal query representations, we use training samples comprising 
a medical image, its associated clinical question, and the corresponding
medical report. Image--question pairs are the input to the
GMU~\cite{arevalo2017gmu}, while the reports serve as supervision in the
Report-Centric configuration and, later, as the contexts of the knowledge base. Visual and textual embeddings, $\mathbf{x}_v$ and $\mathbf{x}_t$, are extracted with frozen encoders under two setups: (i) a dual-encoder with MobileNetV2~\cite{sandler2018mobilenetv2} for images and PubMedBERT~\cite{PubMedBERT} for text; and (ii) BiomedCLIP~\cite{biomedclip}. 
Given these embeddings, the GMU first projects each modality into a shared latent space of dimension $d$:
\begin{equation}
  \mathbf{h}_v = \tanh(\mathbf{W}_v\mathbf{x}_v+\mathbf{b}_v), \qquad
  \mathbf{h}_t = \tanh(\mathbf{W}_t\mathbf{x}_t+\mathbf{b}_t),
\end{equation}
where $\mathbf{W}_v$, $\mathbf{W}_t$, $\mathbf{b}_v$, and $\mathbf{b}_t$ are learnable parameters. A gate then weights the contribution of each modality:
\begin{equation}
  \mathbf{z}=\sigma(\mathbf{W}_z[\mathbf{x}_v;\mathbf{x}_t]+\mathbf{b}_z),
\end{equation}
The joint representation is
\begin{equation}\label{eq:gmu}
\mathbf{h}=\mathbf{z}\odot\mathbf{h}_v+(1-\mathbf{z})\odot\mathbf{h}_t,
\end{equation}
$\mathbf{h}$ is an adaptive,
per-dimension mixture of visual and textual information.
We train the GMU with a contrastive objective that pulls $\mathbf{h}$ toward its positive and pushes it away from negatives, under four configurations that differ in the alignment objective. 
This model maps 
every image-question pair  to its own multimodal embedding $\mathbf{h}$, i.e., 
the contribution of each modality is instance-specific. 
Four variants to selecting positive samples ($\mathbf{k}^{+}$) in a InfoNCE loss~\cite{oord2018representation},  function were designed:
\begin{itemize}
  \item \textbf{Baseline.} The joint
  representation is learned directly from the multimodal fusion, without an additional semantic anchor.
  \item \textbf{Fusion-Image (F-I).} The positive is the embedding of the  corresponding medical image; negatives are non-matching images.
  \item \textbf{Fusion-Question (F-Q).} The positive is the embedding of the associated clinical question; negatives are unrelated questions.
  \item \textbf{Fusion-Report (F-R).} The positive is the embedding of the clinical report associated with the image--question pair; negatives are non-matching reports.
\end{itemize}
\paragraph{2. Knowledge Indexing (per configuration).}
The Knowledge Base is built offline from the medical reports the training data, where each report is a retrievable context. Since the query lives in the joint space learned by the GMU, the contexts must be embedded in that same space for their cosine similarity to be meaningful. Therefore, for each configuration, every report is encoded with the text encoder used for the questions (PubMedBERT or BiomedCLIP), yielding $\mathbf{x}_t$, and passed through the corresponding trained GMU with a zero vector as the visual input ($\mathbf{x}_v=\mathbf{0}$). The resulting $d$-dimensional embeddings are stored in a FAISS~\cite{douze2024faiss} index, so that one index is built per configuration. At inference time, the query is embedded with the same GMU and the top-$k$ most similar contexts are retrieved. 
\paragraph{3. Test Example Encoding.}  At inference time, each example of the test split 
is converted into a multimodal query. The image and the question are encoded with the same frozen encoders used during training, and the resulting
embeddings are fed to the trained GMU, Eq. (\ref{eq:gmu}).
\paragraph{4. Retrieval.} 
Given the multimodal query $\mathbf{q}$ of a test example, we search the
knowledge index of the corresponding configuration by computing its cosine similarity against every context embedding $\mathbf{c}_j$: $s_j=\text{sim}(\mathbf{q},\mathbf{c}_j)$
and retrieve the $k$ contexts with the highest scores, where $k$ determines the number of medical reports provided as additional evidence to the Med-LVLM. Since the amount of retrieved information may influence answer generation, we evaluate different values of $k$ to analyze their impact on performance, using $k=5$ as the default setting. The text of the retrieved reports is then passed to the generation stage.
\paragraph{5. Generation \& Evaluation.}
We use a simple in-context prompt for answering with LLaVA-Med-1.5~\cite{li2023llavamed}, kept frozen (no fine-tuning): the top-$k$ retrieved contexts are combined with the image
and the question in the prompt. 

\section{Experiments and results} 
\subsection{Experimental settings}
The GMU module is trained and evaluated using three publicly available medical imaging datasets: MIMIC-CXR\cite{PhysioNet-mimic-cxr-2.1.0}, IU-XRAY~\cite{demner2016iu}, and Harvard-FairVLMed~\cite{luo2024fairclip}, see 
Table~\ref{tab:datasets}.
\begin{table}[t]
\centering
\caption{Datasets and binary VQA splits used in this work.}
\label{tab:datasets}
\footnotesize{
\begin{tabular}{lccc}
\toprule
\textbf{Dataset} & \textbf{Source data} & \textbf{Train (+/-)} & \textbf{Test} \\
\midrule
MIMIC-CXR~\cite{PhysioNet-mimic-cxr-2.1.0} & \shortstack{377,110 X-rays /\\ 227,835 studies} & 962 / 6,532 & 3,468 \\
IU-XRAY~\cite{demner2016iu} & \shortstack{7,470 X-rays /\\ 3,955 reports} & 1,868 / 6,022 & 2,569 \\
Harvard-FairVLMed~\cite{luo2024fairclip} & 10,000 glaucoma samples & 851 / 6,457 & 4,266 \\
\bottomrule
\end{tabular}}
\end{table}
All experiments were performed in Google Colaboratory Pro using T4, L4, and A100 GPUs. 
We used \textbf{LLaVA-Med-1.5 7B}~\cite{li2023llavamed} as the Med-LVLM. 
We evaluated two encoders: $(i)$ a dual-encoder setup, with MobileNetV2~\cite{sandler2018mobilenetv2} for images and PubMedBERT~\cite{PubMedBERT} for text; and $(ii)$ \textbf{BiomedCLIP}~\cite{biomedclip}, which maps visual and textual inputs into a shared embedding space. 
For retrieval, textual report embeddings were indexed offline using FAISS~\cite{douze2024faiss}.

\subsection{Results}
Table~\ref{resultados1} presents the experimental results in terms of \textit{Accuracy}, \textit{F1-score}, and \textit{AUROC} (0--100 scale), considering two encoder settings and the four GMU training configurations. Fusion-Image, Fusion-Question, and Fusion-Report are abbreviated as \textbf{F-I}, \textbf{F-Q}, and \textbf{F-R}, respectively. We include the following references:
LLaVA-Med-1.5 (the MedLVLM at the inference stage without RAG), 
two unimodal retrieval baselines, in which the query is built only from the clinical question (\textit{text-only}) or only from the medical image (\textit{image-only}) and the retrieval-only configurations of RULE~\cite{xia2024rule} and MMed-RAG~\cite{xia2025mmedrag}.

\textbf{BiomedCLIP outperforms the dual encoder in every configuration of our method}, with average gains of +3.9 to +14.1 points (the largest in AUROC on MIMIC-CXR). Also, the encoder directly influences the contribution of the loss configuration: with the dual encoder, the anchored variants often improve the AUROC of the Baseline, whereas \textbf{with BiomedCLIP the Baseline is already strong}. This may be due to the fact that BiomedCLIP's image and text embeddings are already aligned when fed to the GMU, while the GMU must also bridge the modality gap when using the dual (separate-modalities) encoder. 
After the Baseline configuration, F-R and F-Q perform similarly, while F-I was the configuration with lowest performance, still obtaining AUC above 73. 

Compared with LLaVA-Med-1.5 without retrieval (i.e., same method but no RAG), the GMU variants improve most metrics on MIMIC-CXR and FairVLMed, with gains of up to 27.7 points in Accuracy and 26.5 in AUROC on FairVLMed; on IU-Xray, improvements are mostly limited to AUROC. Improvements are also reported for the baselines implementing \emph{unimodal} RAG (rows 10-11): considerable improvements are observed for FairVLMed, moderate ones for MIMIC-CXR, and lower performance than image-only RAG was obtained for IU-Xray. 
\textbf{These results clearly illustrate the benefits of using the learned multimodal representations for RAG. } Compared with RULE and MMed-RAG, our method obtains mixed results: outperforming the references in some metrics and datasets, and achieving lower performance in others. Although the results are not conclusive, they indicate that \textbf{our method achieves performance comparable to that of these state-of-the-art approaches}. Further investigation is needed to identify the conditions under which our method offers consistent improvements.
\section{Conclusions} 
\label{sec:conclusion}
We introduced a method for learning multimodal representations for supporting RAG in Medical VQA.  A GMU learns to weight the contribution of visual and textual information in the query, enhancing the RAG stage by providing useful information about the interaction of modalities. The experimental evaluation shows the competitiveness of our proposed methodology, yet further work is needed to characterize the configurations under which improvements are consistent. 

\begin{table*}[htb]
\centering
\caption{Results on closed-ended Medical VQA. F-R, F-Q and F-I denote the
GMU configurations. Results marked with $*$
correspond to the retrieval-only configurations of RULE~\cite{xia2024rule} and MMed-RAG~\cite{xia2025mmedrag}, i.e., retrieved contexts added to the Med-LVLM prompt without preference fine-tuning, taken from the original papers. Best results per column are in \textbf{bold}.}
\label{resultados1}
\tiny
\setlength{\tabcolsep}{4pt}
\resizebox{\textwidth}{!}{%
\begin{tabular}{lccccccccc}
\toprule
& \multicolumn{3}{c}{\textbf{IU-Xray}} & \multicolumn{3}{c}{\textbf{MIMIC-CXR}} & \multicolumn{3}{c}{\textbf{FairVLMed}} \\
\cmidrule(lr){2-4} \cmidrule(lr){5-7} \cmidrule(lr){8-10}
\textbf{Method} & Acc & F1 & AUC & Acc & F1 & AUC & Acc & F1 & AUC \\
\midrule
\multicolumn{10}{c}{\textit{GMU with dual encoder}} \\
Baseline & 62.79 & 55.78 & 70.92 & 73.79 & 81.86 & 67.12 & 72.11 & 81.80 & 70.90 \\
F-R      & 66.33 & 59.33 & 74.64 & 74.94 & 81.97 & 69.66 & 72.39 & 82.24 & 68.08 \\
F-Q      & 61.74 & 55.78 & 71.00 & 76.73 & 82.48 & 73.05 & 78.79 & 86.55 & 78.34 \\
F-I      & 59.28 & 54.84 & 70.03 & 76.01 & 82.51 & 71.23 & 87.58 & 92.90 & 70.25 \\
\midrule
\multicolumn{10}{c}{\textit{GMU with BiomedCLIP}} \\
Baseline & 76.06 & 65.31 & 85.56 & 82.61 & 86.41 & \textbf{88.32} & \textbf{90.72} & \textbf{94.80} & 71.26 \\
F-R      & 75.47 & 62.88 & 85.59 & 82.41 & 86.24 & 88.15 & 88.47 & 93.26 & 89.30 \\
F-Q      & 72.75 & 62.88 & 85.42 & 82.41 & 86.52 & 87.62 & 87.22 & 92.40 & \textbf{89.50} \\
F-I      & 62.63 & 73.80 & 73.64 & 79.98 & 85.14 & 73.32 & 90.11 & 94.48 & 73.32 \\
\midrule
\multicolumn{10}{l}{\textit{Baselines and Prior Work}} \\
LLaVA-Med-1.5            & 75.47 & 64.04 & 67.46 & 75.79 & 80.49 & 68.84 & 63.03 & 74.11 & 63.05 \\
Text-only retrieval & 78.45 & 69.00 & 89.31 & 81.90 & 85.77 & 87.24 & 70.46 & 80.71 & 70.72 \\ Image-only retrieval  & 86.83 & 74.12 & \textbf{91.05} & 73.25 & 76.55 & 83.32 & 49.86 & 61.26 & 67.60 \\\hline
RULE$^{*}$               & 87.84 & 78.00 & 85.78 & \textbf{83.92} & 87.49 & 83.44 & 87.12 & 92.89 & 77.08 \\
MMed-RAG$^{*}$           & \textbf{89.54} & \textbf{80.72} & 87.13 & 83.57 & \textbf{88.49} & 85.08 & 87.94 & 92.78 & 80.81 \\
\bottomrule
\end{tabular}}
\end{table*}
Table~\ref{resultados2} shows an ablation study on the value of $k$. According to this table, $k=5$ generally results in  the best performance, while $k=1$ often reduces F1-score and AUROC.  This suggests that retrieving multiple reports provides richer clinical evidence for the frozen Med-LVLM.

\begin{table}[H]
\centering
\caption{Ablation study for the number of retrieved contexts $k$. 
}
\label{resultados2}
\tiny
\setlength{\tabcolsep}{2pt}
\begin{tabular}{lcccccccccc}
\toprule
& & \multicolumn{3}{c}{\textbf{IU-Xray}} & \multicolumn{3}{c}{\textbf{MIMIC-CXR}} & \multicolumn{3}{c}{\textbf{FairVLMed}} \\
\cmidrule(lr){3-5} \cmidrule(lr){6-8} \cmidrule(lr){9-11}
\textbf{Model} & $k$ & Acc & F1 & AUC & Acc & F1 & AUC & Acc & F1 & AUC \\
\midrule
Baseline & 1 & 62.39 & 56.44 & 82.74 & 80.36 & 85.75 & 86.85 & 89.06 & 93.82 & 57.48 \\
         & 3 & 75.16 & 64.79 & 85.45 & \textbf{82.75} & \textbf{86.60} & 88.16 & 90.18 & 94.54 & 68.33 \\
         & 5 & \textbf{76.06} & \textbf{65.31} & \textbf{85.56} & 82.61 & 86.41 & \textbf{88.32} & \textbf{90.72} & \textbf{94.80} & \textbf{71.26} \\
\midrule
F-R & 1 & 62.08 & 56.28 & 82.64 & 80.19 & 85.64 & 86.74 & \textbf{89.45} & \textbf{94.02} & 89.23 \\
    & 3 & 74.81 & \textbf{64.31} & 85.51 & \textbf{82.49} & \textbf{86.37} & 87.95 & 85.93 & 91.55 & 89.20 \\
    & 5 & \textbf{75.47} & 62.88 & \textbf{85.59} & 82.41 & 86.24 & \textbf{88.15} & 88.47 & 93.26 & \textbf{89.30} \\
\midrule
F-Q & 1 & 53.60 & 51.77 & 81.84 & 76.61 & 83.71 & 85.81 & \textbf{88.84} & \textbf{93.85} & 85.82 \\
    & 3 & \textbf{74.42} & \textbf{64.11} & 85.01 & \textbf{82.55} & 86.49 & \textbf{87.64} & 87.15 & 92.40 & 89.17 \\
    & 5 & 72.75 & 62.88 & \textbf{85.42} & 82.41 & \textbf{86.52} & 87.62 & 87.22 & 92.40 & \textbf{89.50} \\
\midrule
F-I & 1 & 59.47 & 70.51 & \textbf{77.97} & 75.05 & 81.27 & 71.16 & 87.62 & 93.29 & 55.64 \\
    & 3 & 60.45 & 71.57 & 74.87 & 75.23 & 81.33 & 71.49 & 89.97 & 94.43 & 67.55 \\
    & 5 & \textbf{62.63} & \textbf{73.80} & 73.64 & \textbf{79.98} & \textbf{85.14} & \textbf{73.32} & \textbf{90.11} & \textbf{94.48} & \textbf{73.32} \\
\bottomrule
\end{tabular}
\end{table}
\bibliographystyle{splncs04}
\bibliography{references}

\end{document}